# 전력의 카본 집중도를 고려한 가변적 기계학습 추론

정지완

싱가포르 아메리칸 스쿨 Jiwanjung0101@gmail.com

# Carbon Intensity-Aware Adaptive Inference of DNNs

Jiwan Jung

Singapore American School

ABSTRACT

DNN inference, known for its significant energy consumption and the resulting high carbon footprint, can be made more sustainable by adapting model size and accuracy to the varying carbon intensity throughout the day. Our heuristic algorithm uses larger, high-accuracy models during low-intensity periods and smaller, lower-accuracy ones during high-intensity periods. We also introduce a metric, carbon-emission efficiency, which quantitatively measures the efficacy of adaptive model selection in terms of carbon footprint. The evaluation showed that the proposed approach could improve the carbon emission efficiency in improving the accuracy of vision recognition services by up to 80%.

## 1. INTRODUCTION

The massive computations involved in the inference of deep neural networks (DNNs) lead to high energy consumption and, consequently, a substantial carbon footprint [1]. This issue has become a critical concern as the demand for DNN applications is growing, and sustainability is becoming an urgent priority in today's climate-conscious world.

Carbon intensity, defined as the amount of $CO_2$ emissions produced per unit of electricity consumed, plays a pivotal role in determining the carbon footprint of DNN inference. Interestingly, this factor is not constant and varies throughout the day [2], primarily due to the changing ratio of renewable energy in the total electricity production. When the share of renewable energy is high, carbon intensity is low, and vice versa. This daily fluctuation of carbon intensity, known as a diurnal pattern, provides a unique opportunity to mitigate the carbon footprint associated with DNN inference.

In this paper, we propose a heuristic approach to address this challenge by selecting the model to performance inference adaptively to the real-time changes in carbon intensity. Our technique leverages smaller, less accurate models during periods of high carbon intensity, and larger, more accurate models when the carbon intensity is low. By doing so, we maintain an optimal balance between the necessary accuracy of DNN inference and the urgency to reduce its carbon footprint. To evaluate the proposed approach, we applied it to the vision recognition inference service.

## 2. BACKGROUND RELATED WORK

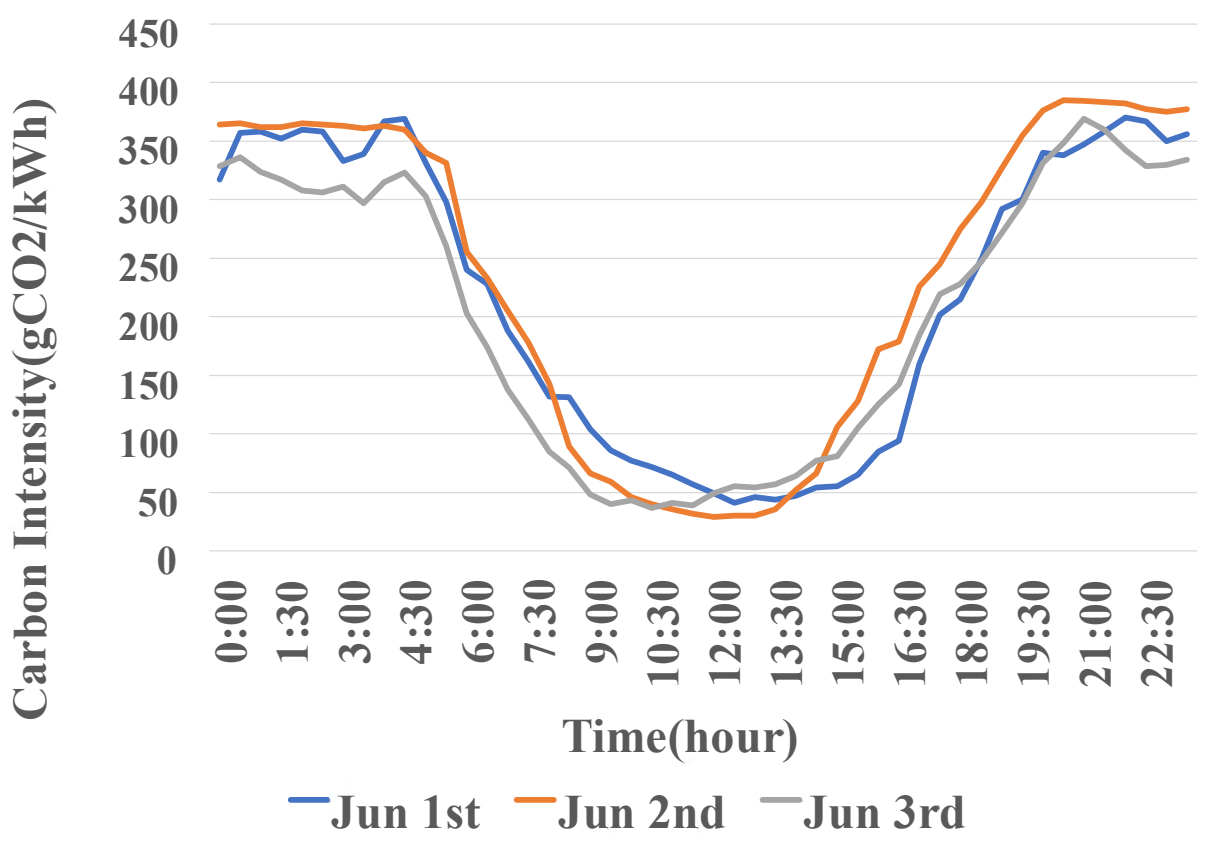

Figure 1 Time Series of Carbon Intensity in South West England from June 1st to 3rd in 2023

Carbon intensity, which is influenced by the proportion of total electricity production derived from renewable energy, exhibits a discernible diurnal pattern, as depicted in Figure 1. This cycle arises from fluctuations in renewable energy production, coupled with the compensatory role of traditional fossil fuel sources during periods of heightened demand. Consequently, daytime typically incurs a surge in electricity demand, leading to elevated carbon intensity.

DNN inference constitutes a substantial segment of cloud ML workloads, accounting for about 70% [5]. These workloads require intensive computational resources, resulting in high energy consumption and, consequently, an enlarged carbon footprint.

Efforts to reduce energy use in DNN inference can directly decrease its carbon footprint. For instance, Nabavinejad et al. [4] developed a scheme adjusting the DNN model's precision and the GPU's DVFS settings in response to server load, aiming for an equilibrium between accuracy, power consumption, and response time. Similarly, Yao et al. [8] examined the efficacy of dynamic voltage and frequency scaling (DVFS) on CNN's energy efficiency, accuracy, and throughput, while proposing a model for balancing latency and energy efficiency. Seo et al. proposed a method that allocates inference requests to various computing accelerators based on their energy efficiencies [6].

Clover reduces carbon emissions from machine learning (ML) inference services using mixed-quality models and GPU partitioning [3]. Our work also employs a mixture of models. However, we demonstrate that a simplified heuristics, which does not require extensive ML techniques, can successfully improve the carbon efficiency of DNN inference. Further, we propose the carbon emission efficiency metric, providing a direct method to quantify carbon efficiency in DNN Inference.

## 3. OUR APPROACH

Table 1 Energy consumption and error rate of representative vision recognition models

| Model | Energy/Inference (mJ/Inference) | Error Rate (%) |
|---|---|---|
| ResNet34 | 359.9321833 | 8.58 |
| ResNet50 | 420.6213298 | 7.138 |
| ResNet101 | 803.0948846 | 6.454 |
| ResNet152 | 1238.147188 | 5.954 |
| VGG16 | 668.9749319 | 9.618 |
| VGG19 | 803.852304 | 9.124 |
| AlexNet | 124.9984724 | 20.934 |

To figure out the correlations between the carbon footprint and the error rate of DNN models, our research begins with a few representative pre-trained vision recognition models, and traces the amount of carbon footprint emitted during their inference. Of course, we could verify that the model with high accuracy involves relatively high carbon footprint per inference, and the model with low accuracy brought out low carbon footprint per inference. To measure the energy consumption, we created a virtual machine instance with NVIDIA V100 GPUs in Google Cloud and set the Triton Inference server in it. The measured results are summarized in Table 1.

As explained in the previous section, the carbon intensity changes with time. In order to obtain the optimal balance between the accuracy and the carbon footprint according to the carbon intensity change, the heuristic algorithm chooses the model to be used for a given request at a certain time point following the equation below.

$$(C_current-C_l)/(C_h-C_l) = (E_Selected-E_l)\ /(E_h-E_l)$$

C_current in the equation is the carbon intensity at the current time, C_low is the lowest carbon intensity among observed values for a certain period of time, and C_high is the highest carbon intensity. Similarly, E_low means the lowest energy consumption per inference among the models, and E_high becomes the highest the energy consumption per inference. Through this formula, E_Selected, the energy consumption per inference of the model suitable for the carbon intensity of the corresponding time period, can be obtained. Among the available models in the model pool, the model with the the energy consumption per inference most similar to this value can be selected as the DNN model to deal with the request that occurred during that time period.

## 4. Evaluation

For evaluation, we utilized the Twitter cache traces [7] as illustrated in Fig. 2. The number of Twitter Request} to emulate the request patterns. In addition, we used the historical records of 30-min. interval carbon intensity changes, which was collected during 2023 June in South West England of the United Kingdom. Finally, we analyzed the energy consumption-per-inference of the eight vision models as shown in Table 1.

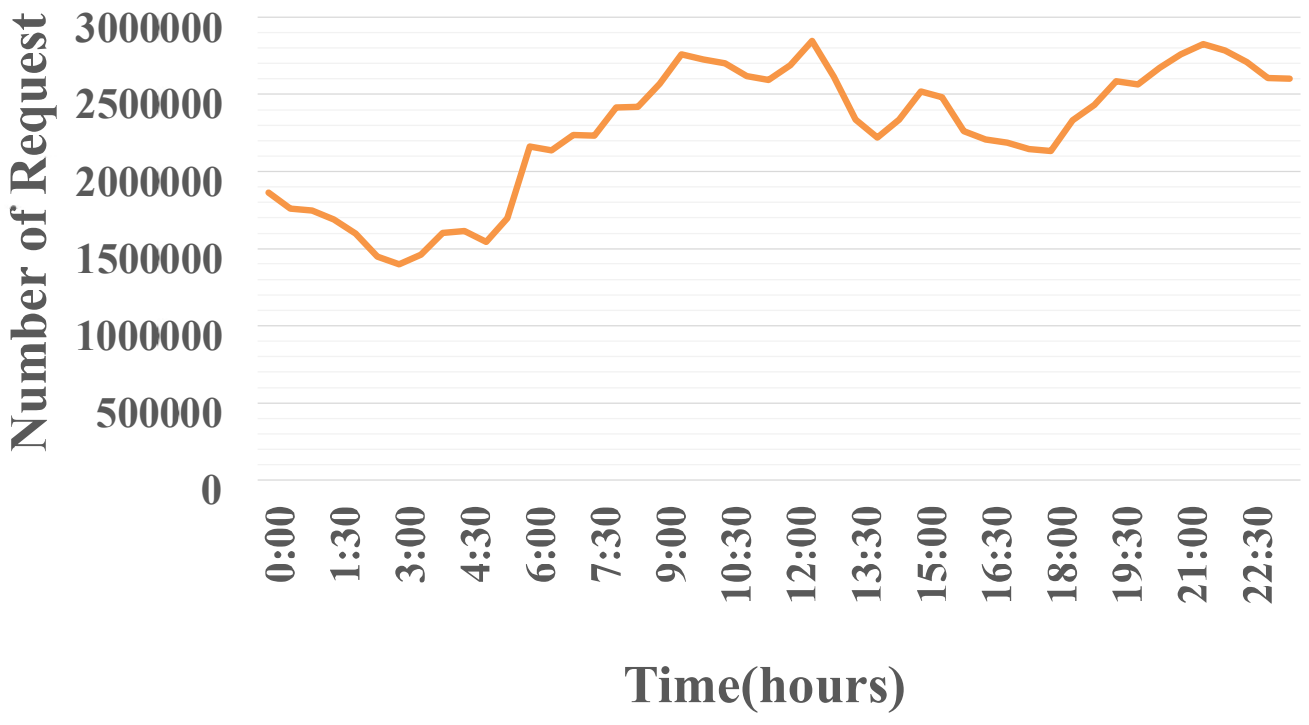

Figure 2 Time Series of Twitter Request Counts

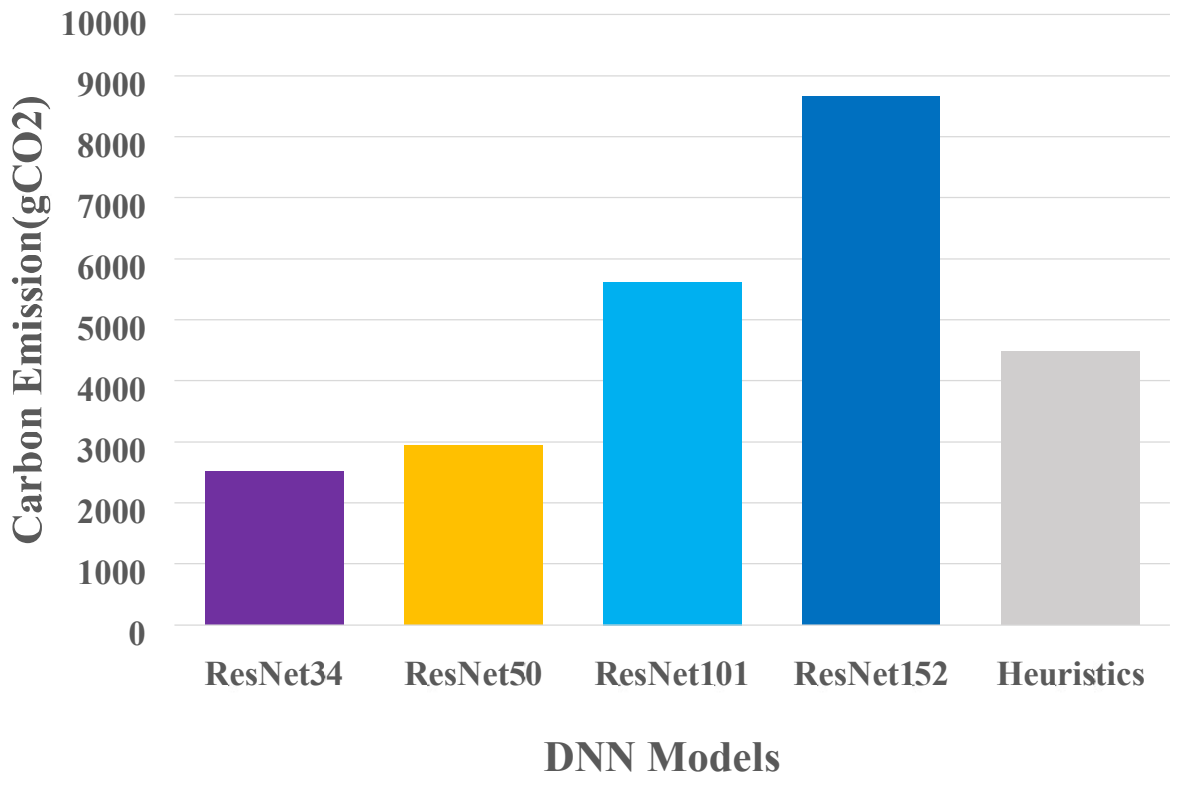

Figure 3 Daily Carbon Emission for Serving Requests

Using these data, we compared the carbon footprint under the proposed scheme with the cases that use a single model. In our evaluation we used only the ResNet variants for their superior accuracy. Fig. 3 shows the results. The error rate of the heuristic approach was 6.57% while that of the cases using a single model can be found in Table 1.

The results show that the heuristic approach significantly reduced carbon production while yielding the similar level of accuracy compared to ResNet101. In addition, when compared with ResNet50, although the carbon footprint was increased, the accuracy was greatly improved. To be specific, based on ResNet50, ResNet101 uses 2676g more carbons to improve the accuracy by 9.58%, while the heuristic approach uses only 1532g of more $CO_2$ to improve the accuracy by 8.00%. Similarly, ResNet152 produced 5720g of additional carbon footprint to improve 16.58% compared to ResNet50.

To put this improvement in perspective, we introduce a metric called carbon emission efficiency. It is defined as the difference in application quality between two configurations, divided by the change in carbon production. This metric quantifies the impact of adaptively adjusting an application's behavior in response to changes in carbon intensity on its carbon footprint.

Our calculations show a carbon emission efficiency of 0.0029 compared to the heuristic's 0.00522, indicating a significant improvement at 16.58% and 6.57% accuracy, with 5720g and 1532g carbon emissions, respectively. These figures confirm that the proposed approach lowered the size of carbon footprint incurred in improving accuracy compared to the case using a predetermined model.

## 5. CONCLUSION AND FUTURE WORKS

This research demonstrated an ability to enhance the carbon emission efficiency of a vision recognition service through a heuristic model selection based on carbon intensity. We plan to apply this approach to inference services across different domains, and apply a non-linear optimization technique such as reinforcement learning to maximize the carbon emission efficiency.